\documentclass{article}

\usepackage{biblatex}
     \usepackage[preprint, nonatbib]{neurips_distshift_2023}

\usepackage{makecell}
\usepackage{changepage}
\usepackage[utf8]{inputenc} %
\usepackage[T1]{fontenc}    %
\usepackage{hyperref}       %
\usepackage{url}            %
\usepackage{booktabs}       %
\usepackage{amsfonts}       %
\usepackage{nicefrac}       %
\usepackage{microtype}      %
\usepackage{xcolor}         %
\usepackage{soul}
\usepackage{amsmath}
\usepackage{graphicx}

\title{Comprehensive OOD Detection Improvements}

\author{%
  Anish Lakkapragada \\
  Booz Allen Hamilton \\
  Annapolis Junction, MD 20701 \\ \texttt{Lakkapragada\_Anish@bah.com} \\
  \And 
  Amol Khanna \\
  Booz Allen Hamilton \\
  Annapolis Junction, MD 20701 \\ \texttt{Khanna\_Amol@bah.com} \\
  \And
  Edward Raff \\
  Booz Allen Hamilton \\
  University of Maryland, Baltimore County \\
  Annapolis Junction, MD 20701 \\ \texttt{Raff\_Edward@bah.com} \\
  \And
  Nathan Inkawhich \\
  Air Force Research Laboratory \\
  Rome, NY 13441 \\ \texttt{Nathan.Inkawhich@us.af.mil} \\
}

\begin{document}

\maketitle

\begin{abstract}
  As machine learning becomes increasingly prevalent in impactful decisions, recognizing when inference data is outside the model's expected input distribution is paramount for giving context to predictions. Out-of-distribution (OOD) detection methods have been created for this task. Such methods can be split into representation-based or logit-based methods from whether they respectively utilize the model's  embeddings or predictions for OOD detection. In contrast to most papers which solely focus on one such group, we address both. We employ dimensionality reduction on feature embeddings in representation-based  methods for both time speedups and improved performance. Additionally, we propose DICE-COL, a modification of the popular logit-based method Directed Sparsification (DICE) that resolves an unnoticed flaw. We demonstrate the effectiveness of our methods on the OpenOODv1.5 benchmark framework, where they significantly improve performance and set state-of-the-art results.
\end{abstract}

\section{Introduction}

Deployed machine learning models today are increasingly involved in  decision systems such as self-driving cars or medical diagnoses. Thus, being able to recognize test data outside a model's typical input distribution is paramount in ensuring that only reliable predictions are utilized. A suite of \emph{out-of-distribution} (OOD) detection methods have been created to detect when data is  beyond the trained input distribution of a model. Such classifiers should also detect  \emph{in-distribution} (ID) data as not OOD during inference.

Many divisions exist between OOD detection methods. A dichotomy we focus on is between methods which utilize the model's feature embeddings (i.e. \emph{representation-based} methods) and those that utilize predictions (i.e. \emph{logit-based} methods). Thus far, most papers have addressed only one type of these methods; in contrast, we contribute to the lineage of both these areas in this paper. 

We explore dimensionality reduction for representation-based methods, which has largely been ignored in current OOD detection methods. We find reducing the dimensionality of the model's representation space significantly improves OOD detection performance. Regarding logit-based methods, we find strong performance, especially on the CIFAR-10 dataset, in our proposed method DICE-COL. DICE-COL is built on top of the popular Directed Sparsification (DICE) logit-based OOD detection method and resolves a flaw in DICE's design.

\section{Related Work}

\subsection{Out-of-Distribution Detection} 

Out-of-Distribution (OOD) detection methods detect when data fed into a supervised machine learning model is beyond its typical input distribution. By convention, OOD detectors provide some scoring function $S(x)$ for a given sample $x$, where a sample is classified as ID if $S(x) \geq \lambda$ and OOD if not. $\lambda$ is a threshold set that such that 95\% of training data is classified as ID.

Because OOD data is not always available, we focus on OOD detection methods that only rely on ID data. Additionally, we focus on \emph{post-hoc} OOD detection methods, which do not interfere in the model's training procedure and thus are greatly preferred. The primary division of OOD detection methods that we comprehensively address is between \emph{logit-based} methods and \emph{representation-based} methods, which differ in the construction of their scoring functions.

\subsection{Logit-Based Methods} 

\emph{Logit-based} methods detect OOD data from the model's predictions. The first popular logit-based method was Maximum Softmax Probability (MSP) \cite{hendrycks2016baseline}, which sets its scoring function to the maximum softmax prediction in the model's output. Energy \cite{liu2020energy}, a related method,  sets its scoring function to the negative of the energy function $E(x)$ of the model's logits: $S(x) = -E(x) = \sum_{i=1}^{C} e^{-f_{i}(x)}$, where $C$ is the number of classes and $f_{i}(x)$ is the model's logit output for the $i$th class. Both methods share the intuition that a sample having only low-confidence predictions is a strong indicator that it is OOD. 

\subparagraph {ASH and DICE}

More recent OOD detection methods \cite{sun2021react, sun2022dice, djurisic2022extremely} apply adjustments to the model exclusively during OOD detection inference  and then utilize the model's predictions for Energy OOD scores. These adjustments typically aim to better separate Energy scores between ID and OOD data \cite{sun2022dice, djurisic2022extremely}. Activation Shaping (ASH) \cite{djurisic2022extremely} during OOD detection inference  sets the lowest 90\% of the flattened representation from the feature extractor to zero. This sparse representation will go onto the neural network’s prediction layer. Directed Sparsification (DICE) \cite{sun2022dice}, a similar method, zeroes the 90\% of the weights of the network's final layer with the lowest contributions to the predictions. Both ASH and DICE have found that penultimate layer activations have an excessively high output variance for OOD samples; this muddles separation between the Energy scores of ID and OOD data. Forcing strict sparsity as ASH and DICE does has been found to alleviate this issue.

\subsection{Representation-Based Methods}

\emph{Representation-based} methods utilize the  network's representations from a given layer (typically the penultimate) for evaluating if a sample is OOD.  Nearest-Neighbors for OOD (abbreviated as KNN) \cite{sun2022out}, a top-performing representation-based method, evaluates if a sample is OOD based on the distance from its penultimate embedding to nearby embeddings of the ID dataset. After training, KNN stores the penultimate embeddings across all or a percentage of the ID dataset. Given a sample with penultimate embedding $z$ at inference, KNN calculates the OOD score as the Euclidean distance of $z$ from its $k$th nearest neighbor in the stored penultimate embeddings, where $k$ is a hyperparameter. 

 Representation-based methods also model distributions on the penultimate layer's representation space. Mahanalobis Distance for OOD (MDS) \cite{lee2018simple} models $C$ class-conditional multivariate gaussian distributions and sets its scoring function to the lowest Mahanalobis Distance between a given sample's penultimate representation and a class's distribution. The mean for each class distribution and the global covariance matrix are modeled with maximum likelihood estimation (MLE).

Relative Mahanalobis Distance (RMDS) \cite{ren2021simple} conducts  the same steps as MDS with a slight modification to MDS's scoring function. After training, RMDS models a multivariate Gaussian "background distribution" of the penultimate layer's representation space for all ID data. The RMDS scoring function is equal to the MDS scoring function minus the Mahanalobis distance between a sample's penulimate representation and the background distribution. RMDS leads to dramatic performance increases relative to MDS \cite{zhang2023openood}.

\subparagraph{Dimensionality Reduction in MDS}

A largely ignored area for OOD detection has been the usage of dimensionality reduction in representation-based methods. On an ID medical image dataset, reducing the dimensionality of the penultimate layer's representation space before applying MDS led to dramatic OOD detection performance improvements \cite{woodland2023dimensionality}.  Distances in higher-dimensional are often less meaningful due to the \emph{curse of dimensionality} \cite{aggarwal2001surprising} and, based on the Manifold hypothesis, many of these higher-dimensional spaces can be accurately represented in fewer dimensions. Note that applying dimensionality reduction here is similar to enforcing sparsity in logit-based methods (e.x. ASH, DICE) in that they both utilize "reduced" features in OOD detection. As an additional bonus, reducing dimensionality leads to faster runtime of OOD detection methods.

\section{Methods}

\subsection{DICE-COL: Post-Hoc Modification of Directed Sparsification (DICE)}

\subparagraph{DICE Introduction} We consider a model with $h(x) \in \mathbb{R}^{d}$ as its penultimate layer output. The weight matrix of this model's final layer is given by $\bold{W} \in \mathbb{R}^{d \times C}$, where $C$ is the number of classes. DICE \cite{sun2022dice} creates a contribution matrix $\bold{V} \in \mathbb{R}^{d \times C}$, where given weight vector $\bold{w}_{c}$ for column $c$ the contribution matrix column $\bold{V}_{c} = \mathbb{E}_{x \in X_{ID}} [\bold{w}_{c} \odot h(x)]$. The $i$th unit for a given column $c$ represents the average contribution of $\bold{W}_{i, c}$ to class $c$. Given hyperparameter $p$ (typically 90\%), DICE sets the lowest $p$\% of entries in $\bold{V}$ to zero and the rest to one to create the mask $\bold{M} \in \mathbb{R}^{d x C}$. DICE replaces the model's weight matrix $\bold{W}$ with $\bold{W} \odot \bold{M}$ for OOD detection  and uses Energy OOD scores.

\subparagraph{DICE-COL} Because DICE calculates the masking matrix $\bold{M}$ across all $\bold{V}$ entries, this means that entire weight columns can be zeroed. This leads to some classes never being predicted. We expect this to hurt OOD detection performance. We propose DICE-COL which calculates mask vectors for each \emph{column} of the weight matrix. DICE-COL's hyperparameter $p$ refers to the percentage of weights in each column to be zeroed.

\subsection{Dimensionality Reduction for Representation-Based Methods}

Considering the aforementioned preliminary success of dimensionality reduction in MDS, we seek to more rigorously test the effectiveness of dimensionality reduction on other representation-based methods. For any given representation-based method, we transform the post-training model's representation space to a lower dimension through PCA. Any other steps of the OOD detection method are performed on this lower-dimensional space. At inference time, we utilize the trained PCA to transform a given sample's penultimate embedding to the lower-dimensional space. From there, the method performs OOD detection as usual.  While other dimensionality reduction methods exist, we use PCA to minimize computational cost. We test dimensionality reduction on representation-based methods MDS, KNN, and RMDS.

\section{Results}

\subsection{Data and Evaluation Setup}

We evaluate our methods on the OpenOODv1.5 benchmark framework \cite{zhang2023openood}, which has evaluated 20 post-hoc methods across 4 ID datasets: CIFAR-10 \cite{krizhevsky2009learning}, CIFAR-100  \cite{krizhevsky2009learning}, ImageNet-1K \cite{deng2009imagenet}, and ImageNet-200, a subset of ImageNet-1K. For each ID dataset, OpenOODv1.5  prescribes OOD datasets \cite{krizhevsky2009learning, le2015tiny, deng2012mnist, netzer2011reading, cimpoi2014describing, zhou2017places, vaze2021open, van2018inaturalist, bitterwolf2023or, wang2022vim} split into \emph{NearOOD} and \emph{FarOOD} datasets based on their visual similarity to the ID dataset.

\begin{table}[h!]
        \caption{Model architecture and Near/FarOOD datasets for each ID dataset we evaluate on.}
        \label{table:benchmarks}
        \centering
        \begin{tabular}{llll}
        \toprule
        ID Dataset & Model &  NearOOD Datasets & FarOOD Datasets  \\
        \midrule
        CIFAR-10 & ResNet-18 & CIFAR-100, TIN  & MNIST, SVHN , Textures, Places365 \\ 
        CIFAR-100 & ResNet-18 & CIFAR-10, TIN & MNIST, SVHN, Textures, Places365 \\ 
        ImageNet-200 & ResNet-18 & SSB-Hard, NINCO & iNaturalist, Textures, OpenImage-O \\ 
        ImageNet-1K & ResNet-50 & SSB-Hard, NINCO & iNaturalist, Textures, OpenImage-O \\ 
        \bottomrule
    \end{tabular}
\end{table}

We use the three model checkpoints for each ID dataset provided by OpenOODv1.5 for fair evaluation. We evaluate our methods based on the AUROC of their OOD detection (binary) predictions.

\subsection{DICE-COL}

\begin{table}[h!]
   \begin{adjustwidth}{-2cm}{}
    \label{table:dicecol}
   \caption{AUROC scores comparing DICE with DICE-COL. For DICE-COL, $p$ is set to 90\% for all datasets.}
    \begin{tabular}{ccccccccc}
        \cmidrule[\heavyrulewidth]{2-9}
         & \multicolumn{2}{c}{CIFAR-10 } & \multicolumn{2}{c}{CIFAR-100} & \multicolumn{2}{c}{ImageNet-200} & \multicolumn{2}{c}{ImageNet-1K} \\ 
       \cmidrule{2-9}
       & NearOOD & FarOOD & NearOOD & FarOOD & NearOOD & FarOOD & NearOOD & FarOOD \\ 
       \midrule
       DICE & \makecell{78.34 ± 0.79}  & \makecell{84.23  ± 1.89} & \makecell{\textbf{79.38 ± 0.23}} & \makecell{80.01  ± 0.18}  & \makecell{\textbf{81.78  ± 0.14}} & \makecell{90.80  ± 0.31} & 73.07 & \textbf{90.95} \\ 
       DICE-COL & \makecell{\textbf{84.15 ± 0.37}} & \makecell{\textbf{88.38 ± 1.57}} & \makecell{79.10 ± 0.27} & \makecell{\textbf{80.06 ± 0.37}} & \makecell{81.73 ± 0.09} & \makecell{90.80 ± 0.34} & \textbf{73.65} & 90.86\\
        \bottomrule
    \end{tabular}
  \end{adjustwidth}
\end{table}

We present our results for DICE-COL in Table \ref{table:dicecol}. We find the biggest benefit to DICE-COL compared to DICE on the CIFAR-10 dataset.

\subsection{Dimensionality Reduction for Representation-Based Methods}

We compare the AUROCs of MDS, RMDS, and KNN with dimensionality reduction to their vanilla methods. A  \hl{highlighted} cell means our method sets a new state-of-the-art record compared to 20 benchmarked post-hoc methods in OpenOODv1.5. We detail our results in the following tables.

\begin{table}[h!]
\begin{adjustwidth}{-2cm}{}
\label{table:knnpca}
\caption{KNN-PCA uses 128 components for all datasets except ImageNet-1K, where 1024 components are used.}
\begin{tabular}{ccccccccc}
    \cmidrule[\heavyrulewidth]{2-9}
    & \multicolumn{2}{c}{CIFAR-10} & \multicolumn{2}{c}{CIFAR-100} & \multicolumn{2}{c}{ImageNet-200} & \multicolumn{2}{c}{ImageNet-1K} \\ 
   \cmidrule{2-9}
   & NearOOD & FarOOD & NearOOD & FarOOD & NearOOD & FarOOD & NearOOD & FarOOD \\ 
    \midrule
   KNN & \makecell{90.64 ± 0.20}  & \makecell{92.96 ± 0.14} & \makecell{80.18 ± 0.15} & \makecell{\textbf{82.40} ± 0.17}  & \makecell{81.57 ± 0.17} & \makecell{\textbf{93.16 ± 0.22}} & \textbf{71.10} & \textbf{90.18} \\ 
   KNN-PCA & \makecell{\hl{\textbf{90.65 ± 0.20}}} & \makecell{92.96 ± 0.14} & \makecell{\textbf{80.28 ± 0.15}} & \makecell{82.25 ± 0.17} & \makecell{\textbf{81.58 ± 0.17}} & \makecell{93.00 ± 0.22} & 68.82 & 85.93 \\
   \bottomrule
\end{tabular}
\end{adjustwidth}
\end{table}

\begin{table}[h!]
\begin{adjustwidth}{-2cm}{}
\label{table:mdspca}
\caption{MDS-PCA uses 128 components for all datasets except ImageNet-1K, where 512 components are used.}
\begin{tabular}{ccccccccc}
    \cmidrule[\heavyrulewidth]{2-9}
    & \multicolumn{2}{c}{CIFAR-10 } & \multicolumn{2}{c}{CIFAR-100} & \multicolumn{2}{c}{ImageNet-200} & \multicolumn{2}{c}{ImageNet-1K} \\ 
   \cmidrule{2-9}
   & NearOOD & FarOOD & NearOOD & FarOOD & NearOOD & FarOOD & NearOOD & FarOOD \\ 
   \midrule
   MDS & \makecell{84.20 ± 2.40}  & \makecell{89.72 ± 1.36} & \makecell{58.69 ± 0.09} & \makecell{69.39 ± 1.39} & \makecell{61.93 ± 0.51} & \makecell{74.72 ± 0.26} & \textbf{55.44} & \textbf{74.25} \\ 
   MDS-PCA & \makecell{\textbf{85.67 ± 2.10}} & \makecell{\textbf{89.96 ±1.45}} & \makecell{\textbf{73.68 ± 0.24}} & \makecell{\textbf{80.39 ± 1.30}} & \makecell{\textbf{73.48 ± 0.41}} & \makecell{\textbf{82.54 ± 0.29}} & 53.78 & 70.79 \\
   \bottomrule
\end{tabular}
\end{adjustwidth}
\end{table}

\begin{table}[h!]
\begin{adjustwidth}{-2cm}{}
\label{table:rmdspca}
\caption{RMDS-PCA uses 128 components for all datasets except ImageNet-1K, where 2048 components are used.}
\begin{tabular}{ccccccccc}
   \cmidrule[\heavyrulewidth]{2-9}
    & \multicolumn{2}{c}{CIFAR-10 } & \multicolumn{2}{c}{CIFAR-100} & \multicolumn{2}{c}{ImageNet-200} & \multicolumn{2}{c}{ImageNet-1K} \\ 
   \cmidrule{2-9}
   & NearOOD & FarOOD & NearOOD & FarOOD & NearOOD & FarOOD & NearOOD & FarOOD \\ 
   \midrule
   RMDS & \makecell{89.80 ± 0.28}  &  \makecell{92.20 ± 0.21} & \makecell{\textbf{80.15 ± 0.11}} & \makecell{82.92 ± 0.42} & \makecell{82.57 ± 0.25} & \makecell{88.06 ± 0.34} & 76.99 & 86.38 \\ 
   RMDS-PCA & \makecell{89.80 ± 0.24} & \makecell{\textbf{92.51 ± 0.37}} & \makecell{80.12 ± 0.08} & \makecell{\textbf{\hl{83.04 ± 0.34}}} & \makecell{\textbf{82.70 ± 0.44}} & \makecell{\textbf{88.61 ± 1.11}} & \textbf{77.08} & \textbf{87.05} \\
   \bottomrule
\end{tabular}
\end{adjustwidth}
\end{table}

We conjecture that the loss of critical dimensions in the lower-dimensional embeddings used in KNN-PCA leads to its mixed results. However, on both KNN and RMDS, state-of-the-art results across the 20 published post-hoc methods have been set. 

We observe the merit of dimensionality reduction applied to representation-based methods in (R)MDS-PCA's results. Through reducing the dimensionality of the feature embeddings, distance calculations performed on them in OOD detection methods are made more meaningful. In addition, these experiments validate the success of distilled features in OOD detection from the novel angle of a representation-based method.

\section{Conclusion}

This paper comprehensively improves representation-based and logit-based methods for OOD detection. We apply dimensionality reduction on representation-based methods, which leads to significant improvements on OOD detection performance and sets state-of-the-art results on the OpenOODv1.5 benchmarks. We propose DICE-COL to improve performance upon popular logit-based method DICE. We find considerable improvements in DICE-COL compared to DICE. We hope this work inspires further research on feature transformations in OOD detection methods.

\printbibliography

\end{document}